\documentclass[10pt,journal,compsoc]{IEEEtran}
\usepackage{graphicx}
\usepackage{amsmath}
\usepackage{color}
\usepackage{amsfonts,amssymb}

\ifCLASSOPTIONcompsoc
  \usepackage[nocompress]{cite}
\else
  \usepackage{cite}
\fi

\ifCLASSINFOpdf

\else

\fi

\begin{document}

\title{Global-to-local Expression-aware Embeddings for Facial Action Unit Detection}

\author{Rudong~An,
        Wei~Zhang,
        Hao~Zeng,
        Wei~Chen,
        Zhigang~Deng,
        Yu~Ding

\IEEEcompsocitemizethanks{\IEEEcompsocthanksitem R. An, W. Zhang, H. Zeng, W, Chen, and Y. Ding are with the Netease Fuxi AI
Lab, Hangzhou, China.\protect\\
\IEEEcompsocthanksitem Z. Deng is with the Department of Computer Science, University of Houston, Houston, Texas, USA.}
\thanks{Manuscript received xxx xx, 2022; revised xxxx xx, 20xx.}}


\IEEEtitleabstractindextext{

\begin{abstract}

Expressions and facial action units (AUs) are two levels of facial behavior descriptors. Expression auxiliary information has been widely used to improve the AU detection performance. However, most existing expression representations utilized in AU detection works can only describe pre-determined discrete categories (e.g., Angry, Disgust, Happy, Sad, etc.) and cannot capture subtle expression transformations like AUs. In this paper, we propose a novel fine-grained \textsl{Global Expression representation Encoder} to capture subtle and continuous facial movements, to promote AU detection. To obtain such a global expression representation, we propose to train an expression embedding model on a large-scale expression dataset according to global expression similarity.
Moreover, considering the local definition of AUs, it is essential to extract local AU features. Therefore, we design a \textsl{Local AU Features Module} to generate local facial features for each AU. Specifically, it consists of an AU feature map extractor and a corresponding AU mask extractor. First, the two extractors transform the global expression representation into AU feature maps and masks, respectively. Then, AU feature maps and their corresponding AU masks are multiplied to generate AU masked features focusing on local facial region. Finally, the AU masked features are fed into an AU classifier for judging the AU occurrence.
Extensive experiment results demonstrate the superiority of our proposed method. Our method validly outperforms previous works and achieves state-of-the-art performances on widely-used face datasets, including BP4D, DISFA, and BP4D+. 
\end{abstract}

}
\maketitle

\IEEEdisplaynontitleabstractindextext

\IEEEpeerreviewmaketitle



\section{Introduction}
\label{sec:introduction}

\begin{figure}[!t]
    \centering
    \includegraphics[ width=9cm]{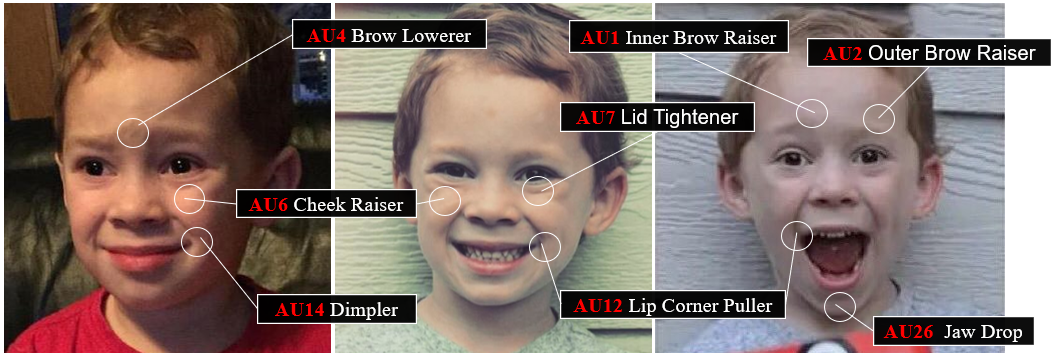}
    \caption{Examples of three expressions and their corresponding AUs. From left to right: doubt, happiness, surprise. Expression is a global description of the face muscle movements, while AUs refer to the individual local description of muscle motion.}
    \label{fig:AU emotion}
\end{figure}

\IEEEPARstart{F}{acial} Affect Analysis (FAA) is an active research area in computer vision and affective computing communities. 
FAA includes two approaches, namely expression and Facial Action Units (AUs). As a universal non-verbal approach to carrying out human communication, facial expressions can be described via a combination of AUs~\cite{FACS}. Facial AUs, coded by the Facial Action Coding System (FACS)~\cite{FACS}, include 32 atomic facial action descriptors based on anatomical facial muscle groups. Table \ref{tab:AUs definition} summarizes the information of the 15 AUs used in this paper, including their names and corresponding involved muscles. Each AU defines the movement of a specific face region. For example, AU1 (Inner Brow Raiser) is a descriptor focused on the medial part of  the frontalis. AUs and expression are highly related to each other. Nearly any possible expression can be described as a specific combination of facial AUs. For instance, as shown in Figure~\ref{fig:AU emotion}, a happy expression can be achieved with the occurrence of both AU6 and AU12; a doubt expression is related to AU4, and a surprise expression is related to AU1 and AU26. From this perspective, AU features are often considered as a kind of facial expression representation.  
In fact, AUs play an important role in conveying human emotions and automatic expression analysis. AU detection has attracted much attention in recent years due to its wide applications, including emotion recognition \cite{AU4EmoctionRecog}, micro-expression detection \cite{AU4MicroExpression}, and mental health diagnosis \cite{AUD4diagnose}. 

AU detection is challenging due to the lack of annotated data and difficulty in capturing local features. Annotating AUs takes hours of tedious effort to manually label one hundred images, even for well-trained experts \cite{zhi2020comprehensive}. Therefore, it is time-consuming and laborious to obtain large scale well-annotated AUs data\cite{peng2018weakly}\cite{niu2019multi}. In terms the lack of labeled data, on the one hand, it often makes the training of deep networks suffer from data over-fitting. On the other hand, the AU datasets usually contain limited identities, dozens\cite{BP4D}\cite{MavadatiDISFA} or hundreds\cite{zhang2016multimodal}, which also leads to identity over-fitting\cite{xiang2017linear}\cite{LP}\cite{chen2022causal}. 
Compared with AU datasets, there are amounts of accessible and easy-to-annotate expression data, like FEC\cite{vemulapalli2019compact} and AffectNet\cite{mollahosseini2017affectnet}, which are beneficial for improving AU detection performance due to their close correlations~\cite{peng2018weakly}\cite{niu2019multi}\cite{xiang2017linear}.  
However, simply regarding expressions as several rough discrete classes is sub-optimal for learning subtle expression transformations \cite{cui2020knowledge}\cite{liu2015inspired}. By contrast, in this paper we measure the subtle distinctions of expressions through similarity learning, leveraging continuous and compact expression features for AU representations. 

As a facial AU is defined anatomically according to the movements of its corresponding facial muscles, it is inherently related to local facial regions, termed as {\it AU's locality}. To this end, how to effectively extract local features from an input face image is a key yet unresolved challenge, because local facial deformations not only are subtle and transient \cite{zhi2020comprehensive}\cite{granger2021weakly} but also vary with individuals \cite{xiang2017linear}\cite{LP}. The straightforward approach is to divide the input image into several local parts in a pre-defined way, which are then fed into networks to capture local information, known as patch learning~\cite{Zhong2015Patch}\cite{liu2014feature}\cite{Onal2019D-PAttNet}. These methods pre-define AU local regions with prior knowledge, cropping a face image into several regular yet coarse-grained patches. Subsequently, the feature extractor is limited to the quality of cropped patches. 

Another category of approaches is to utilize structured geometric information like landmarks to crop more fine-grained local partitions such as Regions Of Interest (ROIs)~\cite{zhao2015joint}\cite{liu2020relation} or attention maps\cite{attnetionTrans2019}\cite{li2018eac}\cite{shao2021jaa}\cite{jacob2021facial}\cite{yang2021exploiting}. 
Some recent works\cite{li2018eac}\cite{shao2021jaa}\cite{jacob2021facial} focus on learning attention maps based on facial landmarks in a supervised manner. The attention maps are usually initialized by pre-defined AU centers based on landmarks and treated as the ground truth during training. 
However, facial AUs occur in specific locations but are not limited to landmark points in most cases~\cite{ma2019r}. Hence, pre-defined AU regions or centers in a uniform way (either by landmarks or prior domain knowledge) are generally error-prone. For example, many prior AU patch generation procedures are fixed for various head poses, while landmark detection tends to deviate under large head rotations, leading to inaccurate AU patches or attention\cite{Onal2019D-PAttNet}.  
In sum, these approaches are sub-optimal in two aspects. First, AU centers based on domain knowledge and landmarks are limited to landmark positions and/or pre-defined rules\cite{ma2019r}\cite{attnetionTrans2019}. Second, landmark detection errors often influence the final performance\cite{ma2019r}\cite{attnetionTrans2019}. Therefore, capturing local features based on manually pre-defined rules with coarse guidances may limit the capability of the local feature extractor. To this end, the sensitive position of each AU in the face feature maps should depend on the AU detection task and the training dataset. In other words, the attention map learning progress should be data-driven. 


Motivated by the above observations, we propose a novel AU detection framework, \textit{Global-to-Local Expression-aware Network(GTLE-Net)}, to address above mentioned challenges. 
Firstly, for the data and identity over-fiting challenge, we pre-train a \textit{global facial expression representation encoder(GEE)} to extract identity-invariant global expression features on a large-scale facial expression dataset, FEC~\cite{vemulapalli2019compact}. Instead of roughly classifying expressions into several discrete categories, we project expressions into a \textit{continuous} representation space, which is more suitable and effective for AU detection in our experiments, since AUs are intrinsically subtle and continuous. Secondly, for the local feature capturing challenge, we design a \textit{local AU feature module (LAM)} with two extractors to produce AU mask and feature map for each AU, respectively. Particularly, the AU mask extractor is learned without any intermediate supervision. 
Then, by multiplying the AU feature maps and the corresponding AU masks, we obtain the \textit{AU masked features}, which filter out irrelevant fields and retain informative ones.
In this way, our framework can not only make full use of large scale well-annotated expression data but also learn attention maps adaptively without extra supervision, to promote AU detection .

To sum up, the main contributions of this paper can be summarized below:
\begin{itemize}
    \item [(1) ]
    A global expression feature encoder(\textit{GEE}) is pre-trained with structured triplet data. This step is proved to be essential for the improvement of AU detection. 
    \item [(2) ]
     An effective attention learning method by a local AU features module (\textit{LAM}) is proposed to capture AU-specific representations, which is crucial to obtain more informative local features and boost up AU detection performance. Visual analytics prove that \textit{LAM} produces highly AU-related local information adaptively.
    \item [(3) ]
    Through extensive experiments, we validate the effectiveness and accuracy of the proposed method on three widely-used benchmark datasets, outperforming the state-of-the-art significantly.  
\end{itemize}

The remainder of this paper is organized as follows. Related works are presented in Section 2. The details of our method are described in Section 3. Comprehensive experiment results are reported in Section 4. Section 5 provides a series of visual analyses. Discussion and concluding remarks are provided in Section 6. Our code will be released if accepted.

\begin{table}[!t]
\footnotesize
\setlength{\tabcolsep}{1.5pt} 
\renewcommand\arraystretch{1.5} 
\centering
{
  \caption{The descriptions and involved muscles of the AUs used in this work. }
  \vspace{-1em}
  \label{tab:AUs definition}
 \begin{tabular}{l|cccccccccccc|c}
    \hline
   \textbf{AU Name} &\textbf{ Description} &\textbf{Involved Facial Muscle(s)} \\[1pt]
    \hline\hline
    AU1   &Inner Brow Raiser &Frontalis, pars medialis     \\
    AU2   &Outer Brow Raiser &Frontalis, pars lateralis       \\
    AU4   &Brow Lowerer     &Depressor Glabellae\\
    AU6   &Cheek Raiser     &Orbicularis oculi \\
    AU7   &Lid Tightener     &Orbicularis oculi\\
    AU9   &Nose Wrinkler     &Levator labii superioris alaquae nasi\\
    AU10  &Upper Lip Raiser  &Levator Labii Superioris \\
    AU12  &Lip Corner Puller &Zygomatic Major  \\
    AU14 &Dimpler &Buccinator\\
    AU15 &Lip Corner Depressor &Depressor anguli oris\\
    AU17 &Chin Raiser &Mentalis\\
    AU23 &Lip Tightener &Orbicularis oris\\
    AU24 &Lip Pressor &Orbicularis oris\\
    AU25 &Lips part &Depressor labii inferioris \\
    AU26 &Jaw Drop &Masseter\\
    \hline
  \end{tabular}}
\end{table}

\section{Related Work}

In recent years, researchers have developed many approaches for AU detection including multi-task and attention mechanisms, introducing some auxiliary information like landmarks, text description, and expressions, etc. We describe previous related works below. 

\subsection{AU Detection with Auxiliary Information}
Due to the high labor cost of AU annotations, the scale and subject variations of AU detection datasets are usually limited. As a result, previous AU detection methods resort to various kinds of auxiliary information to improve the generalization performance.
Facial landmarks are the most widely used pre-trained features for AU detection.
JPML~\cite{JPML} employs facial landmark features to crop local image patches for different AUs.
EAC-Net~\cite{li2018eac} constructs local regions of interest and spatial attention maps from the facial landmarks.
LP-Net~\cite{LP} trains an individual-specific shape regularization network from the detected facial landmarks.
J{\^A}A-Net~\cite{shao2021jaa} jointly performs AU detection and facial landmark detection from the data annotated with both labels.

Other types of auxiliary information have also been explored.
Zhao et al.~\cite{zhao2018learning} pre-train a weakly supervised embedding from a large number of web images.
Cui et al.\cite{cui2020knowledge} summarize the prior probabilities of AU occurrences as generic knowledge. 
Emotions and AUs are jointly optimized under the prior probabilities.
Recently, SEV-Net~\cite{yang2021exploiting} leverages textual descriptions of AU occurrences by employing a pre-trained word embedding to obtain auxiliary textual features.
To enhance the generalization of our model, our method introduces a pre-trained expression embedding as the auxiliary information. 

\subsection{AU Feature Learning}
Due to the local definitions of AUs, it is essential to extract local AU features. As such, some researchers proposed to obtain local information through patch learning. For instance, Zhong et al.~\cite{Zhong2015Patch} and Liu et al.~\cite{liu2014feature} preprocess an input image into uniform patches before encoding to analyze facial expressions. Taking the head pose into consideration, Onal et al.\cite{Onal2019D-PAttNet,Ertugrul-2019-119657} crop AU-specific local facial patches containing information for specific AU recognition after registering the 3D head pose to reduce changes from head movements. Besides, it is a common practice to use attention mechanisms to highlight the features at the facial AU-based positions. 
Facial landmarks with sparse facial geometric features show an advantage in operating as a supervised attention prior. 
EAC-Net~\cite{li2018eac} creates fixed attention maps related to the correlations between AUs and landmarks. 
J{\^A}A-Net~\cite{shao2021jaa} jointly performs AU detection and facial landmark detection, and the predicted landmarks are used to compute the attention map for each AU.
Jacob et al.~\cite{jacob2021facial} propose a multi-task method that combines the tasks of AU detection and landmark-based attention map prediction. 
ARL et al.~\cite{attnetionTrans2019} proposed  a channel-wise and spatial attention learning for each AU. And, a pixel-level relation is learned by CRF to refine the spatial attention. Except for the facial landmarks, SEV-Net~\cite{yang2021exploiting} utilizes the textual descriptions of local details to generate a regional attention map. In this way, it highlights the local parts of global features. However, it requires extra annotations for descriptions. Our work proposes a pixel-wise self-attention map that is data-driven without supervision. Our experiments prove that our attention map is superior to the attention maps produced by prior knowledge.   

\subsection{Expression Representations}
Expressions are essential auxiliary information for AU detection. Some prior works leverage amounts of accessible expression data to enhance AU detection\cite{xiang2017linear}\cite{niu2019multi}\cite{zhao2018learning}\cite{cui2020knowledge}. Recently, Chang et al.\cite{chang2022knowledge} utilize a large amount of unlabeled images to train a representation encoder to extract local representations and project them to a low-dimensional latent space, then improve network performance through contrastive learning. Many methods map face images into a low-dimensional manifold for subject-independent expression representations. 
Early works \cite{mollahosseini2017affectnet}\cite{wang2020region}\cite{zhao2021robust} train the embeddings for discrete emotion classification tasks but neglect the facial expression variations within each class. 
The 3D Morphable Model (3DMM) \cite{paysan20093d}\cite{booth2018lsfm} has been proposed to fit identities and expression parameters from a single face image. 
Expressions are represented as the coefficients of predefined blendshapes in 3DMM.
The estimated expression coefficients are then used for talking head synthesis \cite{li2021write}\cite{zhang2021flow}, expression transfer \cite{yao2021one}\cite{kim2018deep}, and face manipulation \cite{geng20193d}. However, the estimated expression coefficients have weaknesses in representing fine-grained expressions.  
To solve the problem, Vemulapalli and Agarwala~\cite{vemulapalli2019compact} proposed a compact embedding for complicated and subtle facial expressions, where facial expression similarity is defined through triplet annotations. Zhang et al.~\cite{zhang2021learning} proposed a Deviation Learning Network (DLN) to remove the identity information from continuous expression embeddings, and thus achieve more compact and smooth representations.

\section{Proposed Method}
\label{sec:methodology}

In this section, we first briefly introduce the problem definition and then describe our proposed GTLE-Net framework.

\begin{figure*}[!t]
    \centering
    \includegraphics[ width=18cm]{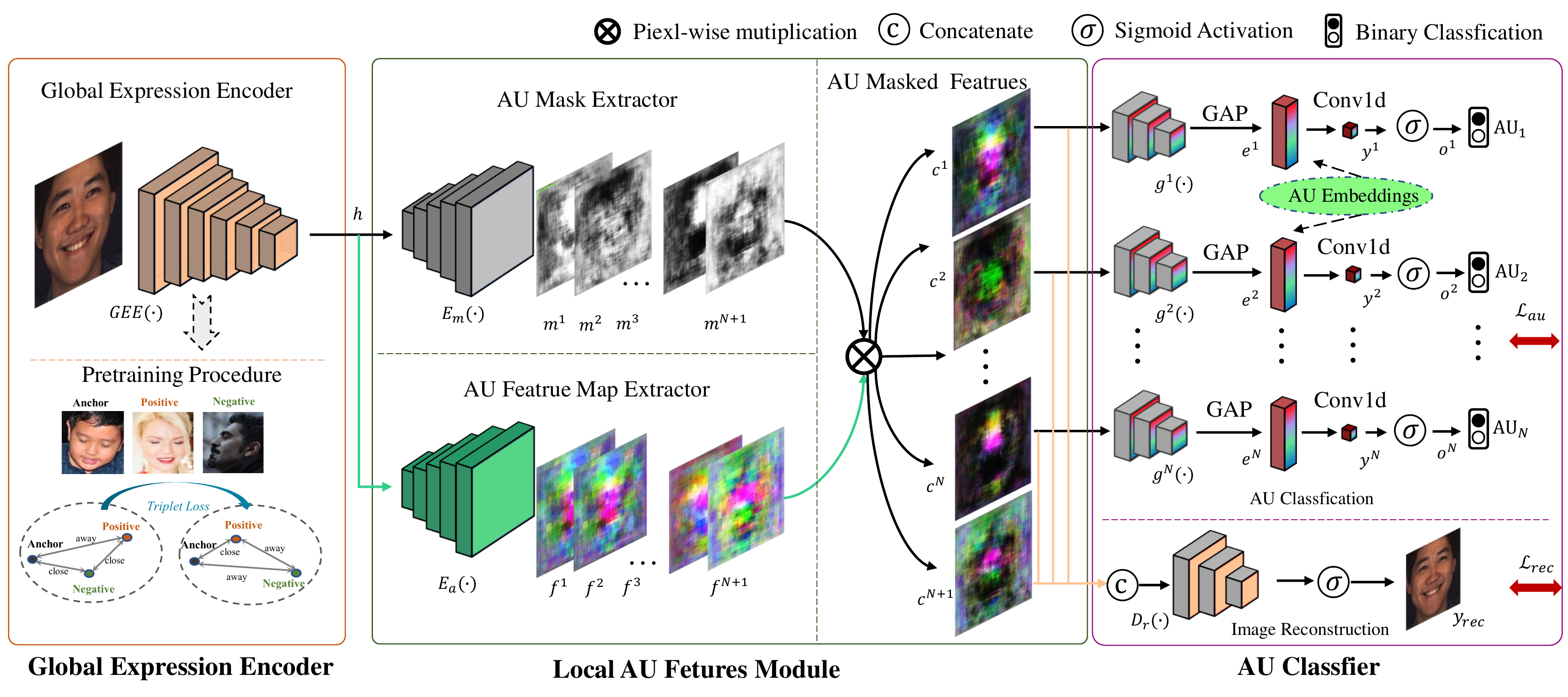}
    \caption{The schematic pipeline of the proposed global-to-local expression-aware embeddings AU detection framework. It is composed of three main modules: the Global Expression representation Encoder(\textsl{GEE}), local AU features module(\textsl{LAM}), and AU classifier. The \textsl{GEE}, pre-trained by expression similarity tasks with the triplet loss, extracts a global expression feature map from the input image, which is capable of sensing subtle expression changes. The global expression feature is then fed into the AU Mask Extractor($E_m$) and the AU Feature Map Extractor($E_a$). The two extractors produce AU-specific masks and feature maps from the global expression feature($h$), respectively. AU masked features are obtained through pixel-wise product on the outputs of two extractors, which are then fed into the following AU Classifier to obtain the AU embeddings and detection results.}
    \label{fig:pipeline}
\end{figure*}

\subsection{Problem Definition}
The task of AU detection is to predict the AU occurrence probabilities $[o^1, o^2,...,o^N]$ given an input face image with a resolution of $256\times256$. 
As described in section 1, expressions and AUs are two means of description from global and local levels respectively. 
They are highly related and thus can be utilized to promote each other. Moreover, facial expression data are easier to obtain and annotate,
while AUs are more subtle and difficult to annotate. Intuitively, facial AU information can be extracted from facial expression representations when expression representation perception is adequately fine-grained. Considering the above factors, we propose a \textit{global expression encoder(GEE)} and pre-train it on a large-scale facial expression dataset, aiming to obtain a powerful and robust expression representation. 
Furthermore, for the purpose of capturing AU local features, we propose a \textit{local AU features module (LAM)} constructed with two extractors, namely, the AU mask extractor ($E_m$) and the AU feature map extractor ($E_a$). 
The framework is illustrated in Fig~\ref{fig:pipeline}. In the following, we discuss each main module in detail.

\subsection{Global Expression Encoder}
\begin{figure}[!t]
    \centering
    \includegraphics[ width=9cm]{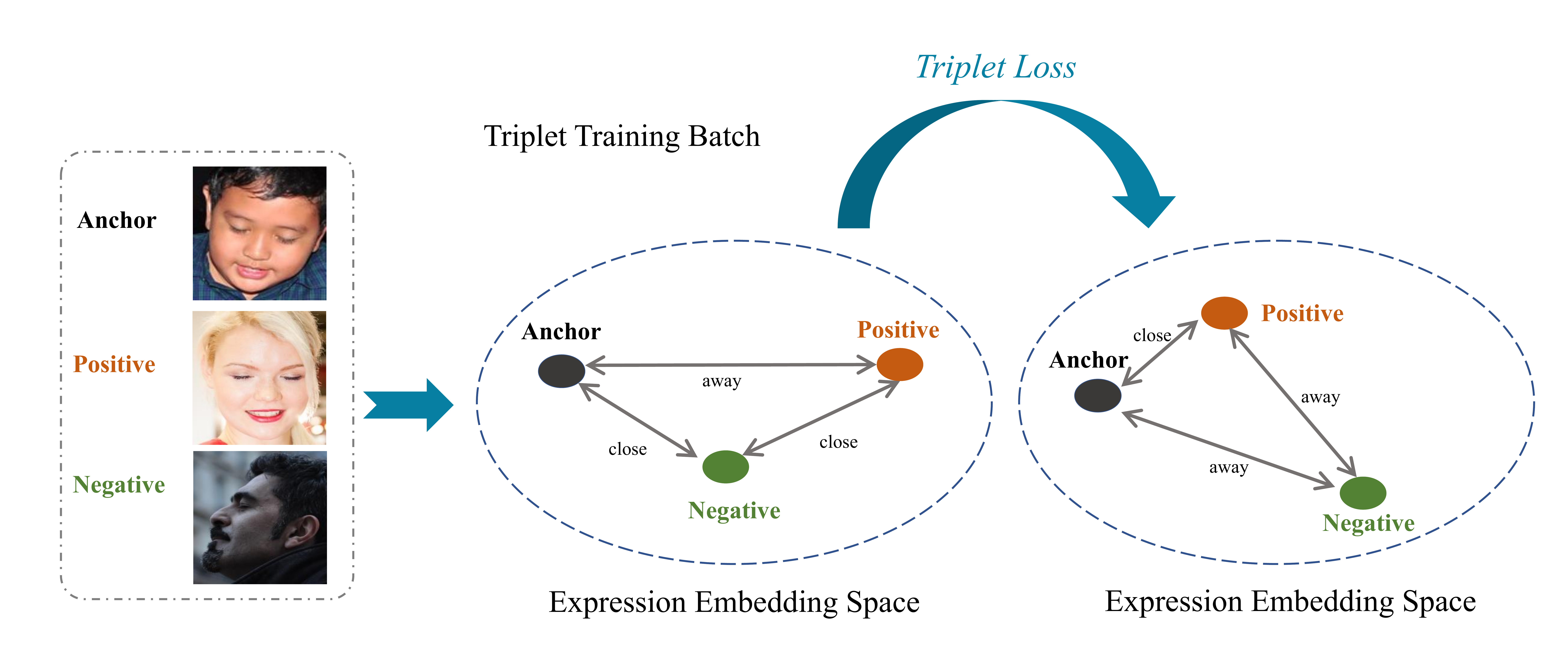}
    \caption{Illustration of our pre-trained expression similarity task. In each triplet, Anchor and Positive have more similar expressions than Negative. By the triplet loss, the pre-trained feature encoder(\textsl{GEE}) minimizes the distances between the Anchor's and Positive's expression embedding and maximizes the distances between the Anchor's and Negative's, Positive's and Negative's expression embedding. }
    \label{fig:triplet_loss}
\end{figure}

To enhance the generalization of our model, we introduce a compact and continuous expression embedding~\cite{vemulapalli2019compact}\cite{zhang2021learning} as the prior auxiliary information. Specifically, \textit{GEE} was pre-trained on the expression similarity task~(See Fig.~\ref{fig:triplet_loss}) using the FEC dataset~\cite{vemulapalli2019compact}. The FEC dataset contains numerous facial image triplets along with the annotations 
indicating which one has the most different expression in each triplet. 
We pre-trained $GEE$ using the triplet loss to constrain that more similar expressions have a smaller embedding distance.
In each triplet, we call the image with the most different expression as the Negative~($N$) and the other two images the Anchor~($A$) and Positive~($P$), respectively. For a triplet~(A, P, N), our $GEE$ outputs three expression embeddings $E_{A}$, $E_{P}$ and $E_{N}$. The pre-trained training process utilizes the triplet loss as follows:
\begin{equation}\small
\begin{aligned}
\mathcal{L}_{tri} &=\max(0,\|E_{A}-E_{P}\|_2^2-\|E_{A}-E_{N}\|_2^2+m) \\
                  &+\max(0,\|E_{A}-E_{P}\|_2^2-\|E_{P}-E_{N}\|_2^2+m) .\label{con: TripletLoss}
\end{aligned}
\end{equation}
Due to the large number of triplets, the pre-trained feature encoder can produce a compact and identity-invariant expression representation, which is beneficial for the downstream expression-related tasks like AU detection.  

We choose the structure of InceptionResnetV1~\cite{szegedy2017inception} as our backbone of $GEE$, replacing the last classifier layer with a linear layer followed by normalization. After the dimensionality reduction, it outputs a 16-dimensional vector as the expression embedding. We first pre-train $GEE$ on the expression similarity task and then choose the shallow layers (up to the layer with feature maps in a resolution of $16\times16$) to extract the global expression representation $h$ from the input face image $im$, which is illustrated as:
\begin{equation}
h = GEE(im)
\end{equation}
Then, our whole framework including $GEE$ is fine-tuned for AU detection. 


Our intuition is twofold. First, $GEE$ initialized from expression embedding learning could instruct the AU detection framework to capture more effective expression features. Second, it obtains a good initialization at the beginning of training, which helps to relieve the problem of over-fitting caused by the limited data and identities of AU datasets. 


\subsection{Local AU Features Module}

The expression features extracted by the pre-trained $GEE$ describe facial movements globally. However, this global representation may not be suitable directly for the subsequent AU detection task, since AUs are defined locally and focus on subtle local motions. Therefore, it is more appropriate to extract local fine-grained features for each AU individually. For this purpose, we design the $LAM$ with two parallel extractors (i.e., a AU mask extractor($E_m$) and an AU feature map extractor($E_a$)) to transform the global expression features into local AU-related representations. 

Specifically, we first extract from the global expression features to produce AU features $f^i$ $(i=1,2,...,N+1)$ with our designed AU feature map extractor $E_a$. Each AU feature is related to a specific AU. However, we found that the AU features generated by the $E_a$ may contain information from unrelated AUs, so we introduce an AU mask extractor($E_m$) to find the region of interest for each AU and also mask out irrelevant content. 

\textbf{AU Feature Map Extractor.} 
The AU feature map extractor $E_a$ aims at generating AU-related features. Given the global expression feature $h$, our AU feature map extractor ($E_a$) performs as follows:
\begin{equation}
[f^1, f^2,...,f^N,f^{N+1}]= E_a(h),
\end{equation}
where $f^k\in\mathbb{R}^{3\times H \times W}$ is the $k$-th AU feature map that is related to the $k$-th AU, and $N$ is the total number of all AUs. $f^{N+1}$ represents the content that is not related to any AUs (e.g., background).

Considering that there are many AUs on the face, if we design a extracting branch for each AU individually, there will be a lot of parameters.
Therefore, we design the AU feature map extractor to be a shared backbone with multiple prediction heads. The shared backbone is formed by stacking multiple residual blocks~\cite{Resnet34} and upsampling layers. The upsampling design can enlarge some local facial details that benefit the subsequent AU detection and reconstruction tasks. Each of the prediction heads consists of a convolutional layer and an activation layer, and it is responsible for predicting the facial content related to a specific AU.

\textbf{AU Mask Extractor.} 
The AU mask extractor $E_m$ is responsible for generating AU masks maps from the global expression features $h$. Each mask represents the region of interest of each AU. Given the global features $h$, the mask maps are predicted by:
\begin{equation}
[m^1, m^2,...,m^N,m^{N+1}]= E_m(h),
\end{equation}
where $m^k \in \mathbb{R}^{1\times H \times W}$ is the $k$-th facial mask for the $k$-th AU, and $m^{N+1}$ is for the background.

To ensure that the values of the predicted facial masks lie in a fixed range of 0 to 1 (0 denotes ``no attention" and 1 denotes ``the strongest attention"), an activation function is added to the last layer of $E_m$. Intuitively, Sigmoid and Softmax are both suitable for this purpose. However, Softmax tend to get sparse attention maps and exclude regions related to the target AU. This may result in the downstream recognition task being unable to learn sufficient information. Hence, we choose Sigmoid to ativate the outputs of $E_m$

\textbf{AU Masked Features.}
Once obtaining the AU specific feature maps and masks, the final AU masked features are generated by multiplying them:
\begin{equation}
[c^1, ...,c^N,c^{N+1}]= [m^1 \times f^1,...,m^N\times f^N,m^{N+1}\times f^{N+1}],
\end{equation}
where $c^i$($i \in (1,2,...,N)$) is the computed masked AU feature for the $i$-th AU that is then used for the subsequent AU detection task. While $c^{N+1}$ is the masked background content which is used for the subsequent face image reconstruction task.

To ensure these AU feature maps contain all facial features of the input image, we introduce a reconstruction task. Specifically, we design an image reconstruction head ($D_r$), which is fed with all masked AU features, and we expect the reconstruction result to be the same as the input facial image. 
\begin{equation}
y_{rec}= D_r([c^1, c^2,...,c^N,c^{N+1}]), 
\end{equation}
where $y_{rec}$ is the reconstruction result.

In this way, the $LAM$ not only realizes the generation of AU local/masked features but also retains the complete information of the input face image. Finally, the generated AU masked features are sent to an AU Classifier (described below) for AU recognition.


\subsection{AU Classifier}

Since the outputs of the $LAM$ are a series of RGB feature maps with a high resolution ($256\times256$), and the number of the feature maps equals the number of AUs in the dataset, the following AU classifier module, denoted as $g^i(\cdot)$, is designed to be a multi-layer CNN architecture, which contains 6 convolution blocks and a global average pooling layer. Each of the first five convolution layers is followed by a batch-norm and an ELU activation function, and their output channels are set to {16, 32, 64, 128, and 256}, respectively. To make sure the sub-module has a wide receptive field, the five CNNs' kernel size, stride, and padding are set to 7, 2, and 3, respectively. The global average pooling layer is set after the fifth convolution block, aiming to aggregate the feature maps into AU Embeddings. In this way, more information closely related to the specific AU will be extracted from the input AU masked feature map. After that, the AU embeddings are fed into the last convolution layers, where all of the output channel, kernel size, stride, and padding are set to 1. This procedure can be formulated as the following function:
\begin{equation}
[y^1, y^2,...,y^N]= [g^1(c^1), g^2(c^2),...,g^N(c^N)].
\end{equation}
Here $y^i$ indicates the output of the last convolution layer of the $i$-th AU. To obtain the final prediction of the AU's probability, we use a Sigmoid activation on the $y^i$ and obtain the predicted probability $o^i$ as follows:
\begin{equation}
[o^1, o^2,...,o^N]= [\sigma(y^1), \sigma(y^2),...,\sigma(y^N)].
\end{equation}

\subsection{Loss Functions}

As illustrated in Figure~\ref{fig:pipeline}, our proposed framework refers to two steps. The pretraining procedure is expression similarity learning, while the AU detection part contains two tasks, namely, AU classification and facial image reconstruction. The pretraining loss ${L}_{tri}$ is a triplet loss, referring to Equation \eqref{con: TripletLoss}, and the AU detection training loss $\mathcal L_{total}$ contains two parts as follows:
\begin{equation}
\mathcal L_{total} = \mathcal L_{au} + \lambda \mathcal L_{rec}.
\end{equation}
where $\mathcal L_{au}$ is the AU classification loss and $\mathcal L_{rec}$ is the facial image reconstruction loss, $\lambda$ is a hyperparameter for adjusting the loss ratio. Our main goal is to recognize the AU, the image reconstruction is merely an auxiliary task, so the $\lambda$ is set to a quite low vale of 0.001 in our experiments. 

AU detection can be conceptually regarded as a multi-label binary classification problem, so the AU detection loss is defined as:
\begin{equation}
\mathcal L_{au} = -\sum_{i=1}^{N} [p_i log \hat{p}_i + (1 - p_i) log(1-\hat{p}_i)],
\end{equation}
where $N$ is the number of AU classes, $p_i$ is the ground truth, and $\hat{p}_i$ is the predicted probability of the $i$-th AU. As for the image reconstruction loss $\mathcal L_{rec}$, we use a pixel-level MSE loss between the input image and the reconstructed one as follows:
\begin{equation}
\mathcal L_{rec} = -\frac{1}{2} \sum_{i=1}^{n}(|| y^i_{rec} - \hat{y}^i_{truth} ||^2_2),
\end{equation}
where $n$ is the pixels number of input image $im$ and equals to $h_{im} \times w_{im}$. Besides, $y^i_{rec}$ and $\hat{y}^i_{truth}$ are the ground-truth and reconstructed value of the $i$-th pixel of the input image, respectively.

\section{Experimental Results}



In this section we report extensive experiments to validate the effectiveness of our method. These experiments are conducted on three public benchmarks commonly-used for AU detection, all of which contain face expression images with manually-annotated AU labels. 

We first introduce the used three datasets in detail and then describe the implementation details of the training and validation processes. Furthermore, we compare our method with the start-of-the-art methods to demonstrate the superiority of our approach. For fair comparisons, our evaluation follows the prior works \cite{li2018eac,shao2021jaa}, using three-fold cross-validation and also their specific division of training and validation data (see below). The reported results are the averages from three-fold experiments. To further obtain insight into our method, we designed ablation experiments to study the impact of the built sub-modules. Finally, we visualize several samples to further validate our method.


\subsection{Datasets}
As mentioned above, our evaluation was performed on three widely-used AU detection datasets: BP4D \cite{BP4D}, DISFA \cite{MavadatiDISFA}, and BP4D+ \cite{zhang2016multimodal}. All the above datasets were annotated with AU labels by certified experts for each frame.

\textbf{BP4D} consists of 328 video clips from 41 participants(23 females and 18 males). Each video frame was annotated with the occurrence or absence of 12 AUs(1, 2, 4, 6, 7, 10,
12, 14, 15, 17, 23, and 24. Their definition can be seen in Table \ref{tab:AUs definition}). In total, about 140,000 frames were annotated frame-by-frame by a group of FACS experts. Following the same subject division \cite{li2018eac,shao2021jaa}, we performed three-fold, subject-exclusive, cross-validations for all experiments.

\begin{table*} 
\footnotesize
\renewcommand\arraystretch{1.5} 
\centering
{
  \caption{Comparisons of our method and the state-of-the-art methods on BP4D in terms of F1 scores(\%). The best results are shown in bold, and the second best results are in brackets.}
  \vspace{-1em}
  \label{tab:F1 on Bp4D}
  \begin{tabular}{l|cccccccccccc|c}
    \hline 
   \textbf{Method} &\textbf{AU1} &\textbf{AU2} &\textbf{AU4}&\textbf{AU6}&\textbf{AU7}&\textbf{AU10}&\textbf{AU12}&\textbf{AU14}&\textbf{AU15} &\textbf{AU17} &\textbf{AU23} &\textbf{AU24} &\textbf{Avg.}\\[1pt]
    \hline \hline 
    LSVM    &23.2 &22.8 &23.1 &27.2 &47.1 &77.2 &63.7 &64.3 &18.4 &33.0 &19.4 &20.7 &35.3 \\
    JPML    &32.6 &25.6 &37.4 &42.3 &50.5 &72.2 &74.1 & 65.7 &38.1 &40.0 &30.4 &42.3 &45.9\\
    DRML    &36.4 &41.8 &43.0 &55.0 &67.0 &66.3 &65.8 &54.1 &33.2 &48.0 &31.7 &30.0 &48.3\\
    EAC-Net &39.0 &35.2 &48.6 &76.1 &72.9 &81.9 &86.2 &58.8 &37.5 &59.1 &35.9 &35.8 &55.9\\
    DSIN    &51.7 &40.4 &56.0 &76.1 &73.5 &79.9 &85.4 &62.7 &37.3 &62.9 &38.8 &41.6 &58.9\\
    ARL     &45.8 &39.8 &55.1 &75.7 &77.2 &82.3 &86.6 &58.8 &47.6 &62.1 &47.4 &55.4 &61.1\\
    LP-Net  &43.4 &38.0 &54.2 &77.1 &76.7 &83.8 &87.2 &63.3 &45.3 &60.5 &48.1 &54.2 &61.0\\
    AU-GCN &46.8 &38.5 &60.1 &[80.1] &79.5 &84.8 &88.0 &67.3 &52.0 &63.2 &40.9 &52.8 &62.8 \\
    SRERL   &46.9 &45.3 &55.6 &77.1 &78.4 &83.5 &87.6 &63.9 &52.2 &63.9 &47.1 &53.3 &62.9\\
    AU-RCNN &50.2 &43.7 &57 &78.5 &78.5 &82.6 &87 &67.7 &49.1 &62.4 &50.4 &49.3 &63.0 \\
    UGN-B   &54.2 &46.4 &56.8 &76.2 &76.7 &82.4 &86.1 &64.7 &51.2 &63.1 &48.5 &53.6 &63.3\\
    J{\^A}A-Net &53.8 &47.8 &58.2 &78.5 &75.8 &82.7 &88.2 &63.7 &43.3 &61.8 &45.6 &49.9 &62.4\\
    SEV-Net &\textbf{58.2} &\textbf{50.4} &58.3 &\textbf{81.9} &73.9 &\textbf{87.8} &87.5 &61.6 &[52.6] &62.2 &44.6 &47.6 &63.9\\
    HMP-PS &53.1 &46.1 &56.0 &76.5 &76.9 &82.1 &86.4 &64.8 &51.5 &63.0 &[49.9] &54.5 &63.4\\
    MHSA-FFN &51.7 &[49.3] &[61.0] &77.8 &\textbf{79.5} &82.9 &86.3 & \textbf{67.6} &51.9 &63.0 &43.7 &[56.3] &64.2\\
    CISNet &54.8 &48.3 &57.2 &76.2 &76.5 &85.2 &87.2 &66.2 &50.9 &[65.0] &47.7 &\textbf{56.5} &64.3\\
    D-PAttNe$t^{tt}$ &50.7 &42.5 &59.0 &79.4 &79.0 &85.0 &[89.3] &67.6 &51.6 &\textbf{65.3} &49.6 &54.5 &64.7\\
    CaFNet &[55.1] &49.3 &57.7 &78.3 &78.6 &[85.1] &86.2 &[67.4] &52.0 &64.4 &48.3 &56.2 &[64.9]\\
    \hline
    \textbf{GTLE-Net}  &\textbf{58.2} &48.7 &\textbf{61.5} &78.7 &[79.2] &84.2 &\textbf{89.8} &66.3 &\textbf{56.7} &64.8 &\textbf{53.5} &53.6 &\textbf{66.3}\\
    \hline
  \end{tabular}}
\end{table*}

\textbf{DISFA} contains 27 video clips from 27 participants(12 females and 15 males). Each video records the facial activity of a participant when watching a 4-minutes video clip. Each clip consists of 4,845 frames evenly. Altogether, around 130,000 frames were annotated with intensities from 0 to 5. Following previous works \cite{jacob2021facial}\cite{li2018eac}\cite{shao2021jaa}\cite{yang2021exploiting}, we regard intensity labels $\{0,1\}$ as absence and others as presence. The evaluation was performed on 8 AUs(1, 2, 4, 6, 9, 12, 25, and 26. Their definition can be seen in Table \ref{tab:AUs definition}). We also performed the same three-fold, subject-exclusive, cross validations on 8 AUs as in the previous works \cite{li2018eac}\cite{shao2021jaa}.

\textbf{BP4D+} contains 1400 video clips from 140 participants(82 females and 58 males). Compared to BP4D, it has larger identity variations and video numbers. About 198,000 frames were annotated with the same 12 AUs in BP4D. We performed three-fold, subject-exclusive, cross-validations to compare our method with the state-of-the-art. Besides, to evaluate the generalization performance, our method was trained on BP4D and then evaluated on BP4D+, following the previous works \cite{attnetionTrans2019}\cite{shao2021jaa}.

To sum up, the order from largest to smallest of the three datasets is BP4D+, BP4D and DISFA. BP4D and BP4D+ were recorded in similar shooting environments and lighting conditions. Both of them share the same 12 AU classes. In comparison, DISFA is the smallest dataset containing only 8 AU classes and was captured in quite different lighting conditions.



\begin{table}[t]
\footnotesize
\setlength{\tabcolsep}{0.9pt} 
\renewcommand\arraystretch{1.5} 
\centering
{ 
  \caption{Comparisons of our method and the state-of-the-art methods on DISFA in terms of F1 scores(\%). 
  }
  \vspace{-1em}
  \label{tab:F1 on DISFA}
  \begin{tabular}{l|cccccccc|c}
   \hline 
 \textbf{Method} &\textbf{AU1} &\textbf{AU2} &\textbf{AU4} &\textbf{AU6} &\textbf{AU9} &\textbf{AU12} &\textbf{AU25} &\textbf{AU26} &\textbf{Avg.}\\[1pt]
    \hline\hline
    LSVM       &10.8 &10.0 &21.8 &15.7 &11.5 &70.4 &12.0 &22.1 &21.8\\
    DRML       &17.3 &17.7 &37.4 &29.0 &10.7 &37.7 &38.5 &20.1 &26.7\\
    APL        &11.4 &12.0 &30.1 &12.4 &10.1 &65.9 &21.4 &26.9 &23.8\\
    EAC-Net        &41.5 &26.4 &66.4 &50.7 &8.5 &\textbf{89.3} &88.9 &15.6 &48.5\\
    DSIN       &42.4 &39.0 &68.4 &28.6 &46.8 &70.8 &90.4 &42.2 &53.6\\
    ARL        &43.9 &42.1 &63.6 &41.8 &40.0 &76.2 &[95.2] &66.8 &58.7\\
    LP-Net     &29.9 &24.7 &72.7 &46.8 &49.6 &72.9 &93.8 &56.0 &56.9\\
    AU-GCN      &32.3 &19.5 &55.7 &57.9 &61.4 &62.7 &90.9 &60.0 &55.0\\
    AU-RCNN  &32.1 &25.9 &59.8 &55.4 &39.8 &67.7 &77.4 &52.6 &51.3 \\
    SRERL      &45.7 &47.8 &56.9 &47.1 &45.6 &73.5 &84.3 &43.6 &55.9\\
    UGN-B      &43.3 &48.1 &63.4 &49.5 &48.2 &72.9 &90.8 &59.0 &60.0\\
    J{\^A}A-Net    &[62.4] &[60.7] &67.1 &41.1 &45.1 &73.5 &90.9 &[67.4] &63.5\\
    MHSA-FFN   &46.1 &48.6 &72.8 &\textbf{56.7} &50.0 &72.1 &90.8 &55.4 &61.5\\
    SEV-Net    &55.3 &53.1 &61.5 &[53.6] &38.2 &71.6 &\textbf{95.7} &41.5 &58.8\\
    HMP-PS     &38.0 &45.9 &65.2 &50.9 &50.8 &76.0 &93.3 &\textbf{67.6} &61.0\\
    CISNet     &48.8 &50.4 &[78.9] &51.9 &47.1 &[80.1] &[95.4] &65.0 &64.7\\
    CaFNet     &45.6 &55.7 &\textbf{80.2} &51.0 &\textbf{54.7} &79.0 &95.2 &65.3 &[65.8]\\
    \hline
    \textbf{GTLE-Net}    &\textbf{63.5} &\textbf{62.5} &69.8 &48.6 &[53.9] &75.6 &94.7 &64.4 &\textbf{66.6}\\
    \hline
  \end{tabular}}
\end{table}

\begin{table*}[!t]
\footnotesize
\renewcommand\arraystretch{1.5} 
\centering
{
  \caption{Comparisons of our method and the state-of-the-art methods on BP4D+ in terms of F1 scores(\%). 
  }
 \vspace{-1em}
  \label{tab:F1 on ThreeFoldBp4DPlus}
  \begin{tabular}{l|cccccccccccc|c}
    \hline
   \textbf{Method} &\textbf{AU1} &\textbf{AU2} &\textbf{AU4} &\textbf{AU6} &\textbf{AU7} &\textbf{AU10} &\textbf{AU12} &\textbf{AU14} &\textbf{AU15} &\textbf{AU17} &\textbf{AU23} &\textbf{AU24} &\textbf{Avg.}\\[1pt]
    \hline \hline
    FACS3D-Net     &43.0 &38.1 &\textbf{49.9} &82.3 &85.1 &87.2 &87.5 &66.0 &48.4 &47.4 &50.0 &[31.9] &59.7\\
    ML-GCN &40.2 &36.9 &32.5 &84.8 &88.9 &89.6 &[89.3] &81.2 &[53.3] &43.1 &55.9 &28.3 &60.3\\
    MS-CAM &38.3 &37.6 &25.0 &85.0 &\textbf{90.9} &\textbf{90.9} &89.0 &81.5 &\textbf{60.9} &40.6 &58.2 &28.0 &60.5\\
    SEV-Net &[47.9] &\textbf{40.8} &31.2 &\textbf{86.9} &87.5 &89.7 &88.9 &\textbf{82.6} &39.9 & \textbf{55.6} &[59.4] &27.1 &[61.5]\\
    \hline
    \textbf{GTLE-Net(Ours)}  &\textbf{51.5} &[46.6] &[43.5] &[86.8] &[89.6] &[91.0] &\textbf{89.8} &[82.3] &46.8 & [49.3] &\textbf{60.9} &\textbf{50.9} &\textbf{65.7}\\
    \hline
  \end{tabular}}
\end{table*}

\begin{table*}[!t]
\footnotesize
\renewcommand\arraystretch{1.5} 
\centering
{
  \caption{Comparisons of our method and the state-of-the-art on cross-dataset evaluation(trained on BP4D and evaluated on BP4D+) in terms of F1 scores(\%). 
  }
 \vspace{-1em}
  \label{tab:F1 on Bp4DPlus}
  \begin{tabular}{l|cccccccccccc|c}
    \hline
   \textbf{Method} &\textbf{AU1} &\textbf{AU2} &\textbf{AU4} &\textbf{AU6} &\textbf{AU7} &\textbf{AU10} &\textbf{AU12} &\textbf{AU14} &\textbf{AU15} &\textbf{AU17} &\textbf{AU23} &\textbf{AU24} &\textbf{Avg.}\\[1pt]
    \hline \hline
    EAC-Net     &38.0 &[37.5] &[32.6] &82.0 &83.4 &87.1 &85.1 &62.1 &44.5 &43.6 &45.0 &32.8 &56.1\\
    ARL         &29.9 &33.1 &27.1 &81.5 &83.0 &84.8 &86.2 &59.7 &[44.6] &43.7 &48.8 &32.3 &54.6\\
    J{\^A}A-Net &\textbf{39.7} &35.6 &30.7 &[82.4] &[84.7] &[88.8] &[87.0] &[62.2] &38.9 &[46.4] &[48.9] &\textbf{36.6} &56.8\\
    \hline
    \textbf{GTLE-Net(Ours)}  &[39.5] &\textbf{37.7} &\textbf{44.0} &\textbf{84.5} &\textbf{84.9} &\textbf{89.6} &\textbf{87.7} &\textbf{72.3} &\textbf{44.9} & [45.3] &\textbf{52.6} &[33.9] &\textbf{59.7}\\
    \hline
  \end{tabular}}
\end{table*}
\begin{table*}[!t]
\footnotesize
\renewcommand\arraystretch{1.5} 
\centering
{
  \caption{Ablation study on BP4D measured by F1 scores(\%). }
  \vspace{-1em}
  \label{tab:F1 on Ablation 2}
 \begin{tabular}{l|cccccccccccc|c}
    \hline
   \textbf{Method} &\textbf{AU1} &\textbf{AU2} &\textbf{AU4} &\textbf{AU6} &\textbf{AU7} &\textbf{AU10} &\textbf{AU12} &\textbf{AU14} &\textbf{AU15} &\textbf{AU17} &\textbf{AU23} &\textbf{AU24} &\textbf{Avg.}\\[1pt]
    \hline\hline
    \textsl{w/o pretrain} &34.9 &24.2 &36.7 &76.1 &71.4 &81.0 &86.3 &58.7 &25.9 &51.9 &27.4 &31.6 &50.5\\
    \textsl{w. GEE Emotion} &48.1 &41.6 &53.9 &78.6 &77.9 &83.2 &88.8 &60.5 &39.3 &65.5 &38.0 &47.6 &60.3 \\
    \textsl{w. GEE ImageNet} &49.6 &39.7 &60.5 &77.3 &78.7 &83.1 &88.6 &[64.6] &48.2 &63.2 &47.5 &48.0 &62.4 \\
    \textsl{w. GEE fixed} &52.3 &45.8 &59.8 &[79.4] &78.1 &\textbf{84.8} &89.0 &64.0 &46.5 &64.0 &38.6 &49.8 &63.5\\
    \hline
    \textsl{w/o $E_m$\&$E_a$} &[55.4] &\textbf{65.1} &\textbf{73.9} &49.0 &46.9 &77.1 &\textbf{92.9} &54.1 &48.1 &64.7 &[52.5] &47.5 &63.3\\
    \textsl{w/o $E_m$}  &53.3 &44.2 &58.7 &78.6 &76.9 &83.9 &[89.8] &62.2 &48.9 &65.1 &[52.5] &51.8 &63.8\\
    \textsl{w/o $E_m$\& w. CNN} &51.8 &43.0 &59.1 &78.3 &78.1 &84.2 &88.7 &63.1 &50.2 &64.0 &49.2 &\textbf{55.1} &63.7\\
    
  
    \hline
    \textsl{\textbf{GTLE-Net(Full)}} &\textbf{58.2} &[48.7] &[61.5] &78.7 &[79.2] &84.2 &[89.8] &\textbf{66.3} &\textbf{56.7} &64.8 &\textbf{53.5} &[53.6] &\textbf{66.3}\\
    \hline
  \end{tabular}}
\end{table*}
\subsection{Implementation Details}

Face images were first aligned and cropped according to landmark detection results. Before feeding to our networks, the input images were resized to $256\times256$. The resolution of feature maps was reduced to $16\times16$ after passing through the $GEE$.

To enhance the feature extraction capability of low-level convolution layers, the $GEE$ was initialized with the corresponding network parameters pretrained on the dataset of ImageNet or FEC, and the rest of our networks were set to Kaiming initialization \cite{he2015delving}. Except for the AU Mask Extractor($E_m$), the whole model parameters were updated by employing an Adam optimizer with hyper-parameters $\beta = 0.9$ and weight decay $10^{-6}$.

The networks were trained for 20 epochs with an initial learning rate of $10^{-5}$ on each dataset fold. The mini-batch size was set to 10. We used a warm-up stage with $10^3$ steps and then decayed the learning rate exponentially with an exponent $-0.5$. The training and inference pipelines were implemented in Pytorch on 4 GeForce GTX 2080ti GPU. 

\subsection{Comparisons with the State-of-the-Art}
We compared our method with many state-of-the-art AU detection methods, including LSVM \cite{LSVM}, JPML \cite{JPML}, DRML \cite{DRML}, APL \cite{LP48}, EAC-Net \cite{li2018eac}, DSIN \cite{LP8}, ARL \cite{attnetionTrans2019}, SRERL \cite{SRERL}, AU-GCN\cite{liu2020relation},AU-RCNN\cite{ma2019r}, ML-GCN \cite{chen2019multi}, MS-CAM \cite{you2020cross}, LP-Net \cite{LP}, FACS3D-Net \cite{yang2019facs3d}, UGN-B \cite{song2021uncertain}, J{\^A}A-Net \cite{shao2021jaa}, MHSA-FFN \cite{jacob2021facial}, SEV-Net \cite{yang2021exploiting}, HMP-PS \cite{song2021hybrid}, D-PAttNe$t^{tt}$\cite{Onal2019D-PAttNet}, CISNet\cite{chen2022causal} and CaFNet\cite{chen2021cafgraph}.

Specifically, while EAC-Net, ARL, J{\^A}A-Net, and SEV-Net were evaluated on all three datasets, LSVM, JPML, DRML, APL, DSIN, SRERL, AU-GCN, AU-RCNN, LP-Net, UGN-B, MHSA-FFN, HMP-PS, CISNet and CaFNet were evaluated on both BP4D and DISFA. Particularly, D-PAttNe$t^{tt}$ was only evaluated on BP4D. FACS3D-Net, ML-GCN and MS-CAM were evaluated merely on BP4D+. Since BP4D+ is released later, it is not used by all the methods.

Following the experimental settings of most previous works, we also employed the F1 score \cite{F1} on the frame level as the evaluation metric. F1-score is a harmonic mean of precision(P) and recall(R), which is defined as $F1 = 2PR/(P+R)$. Since F1-score takes both false positives and false negatives into account, it reflects a more rational measurement of the model's performance than accuracy, especially on datasets with uneven class distributions like BP4D, DISFA and BP4D+.

\textbf{Evaluation on BP4D.}
The results on BP4D are shown in Table \ref{tab:F1 on Bp4D}. The average F1 score of our method is 66.3\%, which significantly outperforms all the other comparison methods, improving about 1.4\% over the second highest one, CaFNet. In terms of the performances on single AUs, our method achieves the highest F1 scores among all the methods on 5 AUs(i.e., AU1, AU4, AU12, AU15, and AU23) and second highest on 1 AU(AU7) of the 12 evaluated AUs. Besides, our method shows over 3.5\% superiority on AU15 and AU23 over the second highest ones.

In terms of AU feature learning, our method outperforms JMPL, DRML, and D-PAttNe$t^{tt}$ about 20.4\%, 18\% and 1.6\%, demonstrating that extracting local features by attention is more effective and our method is superior in AU feature extraction than patch learning. 

Compared with EAC-Net, JAA-Net and MHSA-FFN, which utilize attention mechanisms with the supervisions of pre-defined AU centers, our method achieves improvements of 10.4\%, 5.2\% and 3.9\% over them, respectively. This validates the superiority of our non-predetermined AU attention learning approach.  


\textbf{Evaluation on DISFA.}
Table \ref{tab:F1 on DISFA} shows the evaluation results on DISFA. As shown in this table, our method achieves the average F1 score of 66.6\%, outperforming all the compared state-of-the-art methods. 

Compared to the most recent works SEV-Net, HMP-PS, MHSA-FFN, CISNet and CaFNet, our method achieves an improvement of 7.8\%, 5.6\%, 5.1\%, 1.9\% and 0.8\% over them, respectively. Unlike EAC-Net and J{\^A}A-Net which are supervised by both AU and landmark labels, our method only utilizes AU labels, but still obtains significant improvements of 18.1\% and 3.1\% over EAC-Net and J{\^A}A-Net, respectively, demonstrating the effectiveness of our method in AU feature and attention learning. Note that our method does not explicitly model AU relationships, but significantly exceeds all relation-based methods including JMPL, DSIN, ARL, AU-GCN and AU-RCNN.


\begin{figure*}[!t]
    \centering
    \includegraphics[ width=18cm]{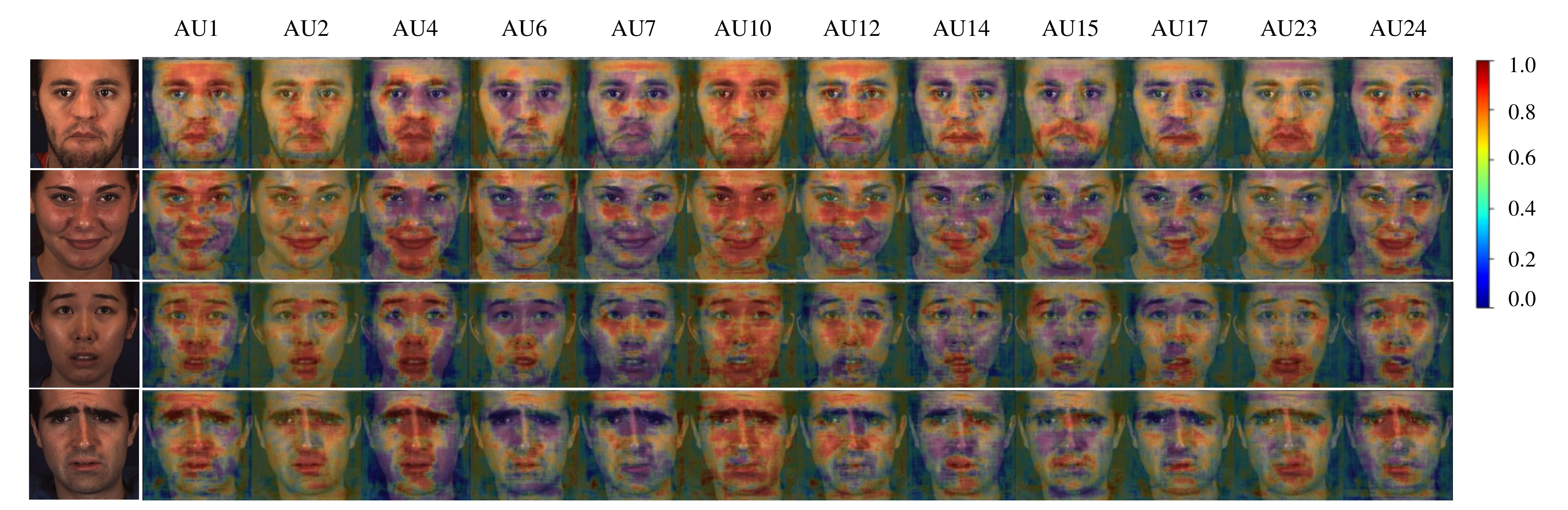}
    \caption{Visualization of the mask attention maps overlaid onto the corresponding input images in BP4D. The first column shows face images with different expressions. We can observe that similar AU-related local mask distributions can be learned by our framework even though the identities and expressions vary significantly.  }
    \label{fig:mask visualization}
\end{figure*}
\begin{figure}[!t]
    \centering
    \includegraphics[ width=9cm]{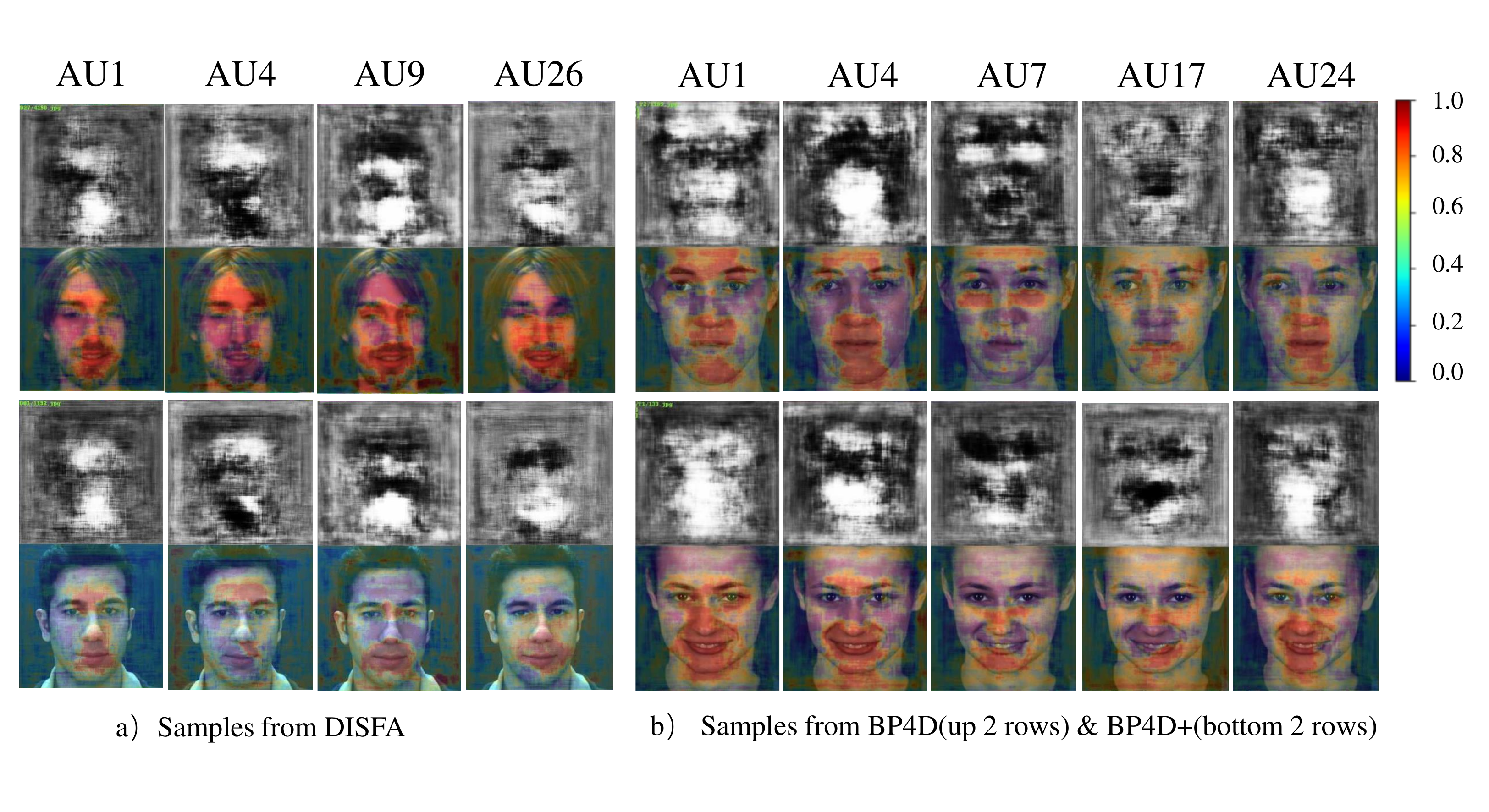}
    \caption{Visualization of the produced mask attention maps:(Left) samples from DISFA, and(Right) samples from BP4D and BP4D+. The learned masks(odd rows) are also visualized and overlaid on the corresponding face images(even rows). The colors covering on the face images from blue to red indicate the attention values from low to high.
    }
    \label{fig:figure1}
\end{figure}

\textbf{Evaluation on BP4D+.}
Table \ref{tab:F1 on ThreeFoldBp4DPlus} shows the evaluation results of the three-fold cross-validation on BP4D+.
In particular, the results of SEV-Net, ML-DCN and MS-CAM are reported in\cite{yang2021exploiting} while the results of EAC-Net are reported in \cite{shao2021jaa}. 
Our method obtains the highest or the second highest results on 11 of the 12 evaluated AUs. It achieves an average F1 score of 65.7\% and outperforms all the state-of-the-art methods, exceeding the second highest method(SEV-Net) over 4.2\%.


To further evaluate the generalization performance of our method on large-scale test identities, we trained our method on BP4D and evaluated it on the full BP4D+ of 140 subjects. The results are shown in Table \ref{tab:F1 on Bp4DPlus}. We use the reported results in the work \cite{shao2021jaa} for comparison. Again, our method outperforms all the compared methods under the large-scale cross-dataset evaluation. It indicates that our method can extract identity-independent information and generalize well to new identities.



\subsection{Ablation Study}

In this section, we conduct ablation studies to evaluate the effectiveness of each key sub-module in our framework, including: 1) the prior knowledge of expression similarity learning, and 2) the local AU features module. In particular, all the ablation experiments are conducted on the BP4D dataset due to space limitations. The experimental results are reported in Table~\ref{tab:F1 on Ablation 2}. 

\textbf{Prior Knowledge.} 
Our framework proposes a vision feature encoder $GEE$ that is pretrained on the FEC dataset via expression similarity learning, which refers to our prior knowledge. To test the efficiency of prior knowledge, we design ablation experiments with different approaches to initialize $GEE$ in the training of \textsl{GTLE-Net}, including:
\begin{itemize}
    \item [a) ]
    \textsl{w/o~pretrain}: $GEE$ is randomly initialized;
    \item [b) ]
    \textsl{w.GEE Emotion}: $GEE$ is initialized via a pretrained classifier of seven discrete facial expression emotion (e.g. neutral, happy, sad, surprise, fear, disgust, angry) categories trained on the AffectNet dataset;
    \item [c) ]
     \textsl{w.~GEE~ImageNet}: $GEE$ is initialized via a pretrained image object classifier built on imageNet \cite{krizhevsky2012imagenet};  
    \item [d) ]
     \textsl{GTLE-Net(Full)}: our proposed method, $GEE$ is  initialized via the facial expression similarity learning; 
    \item [e) ]
     \textsl{w.~GEE~fixed}: $GEE$ is parameterized by the facial expression similarity approach but not updated in the training of \textsl{GTLE-Net}.
\end{itemize}


Firstly, in comparison with \textsl{w/o pretrain},
\textsl{w.GEE Emotion}, \textsl{w. GEE ImageNet} and \textsl{w.~GEE~fixed} achieve improvements of 9.8\%(from 50.5\% to 60.3\%), 11.9\%(from 50.5\% to 62.4\%) and 15.8\%(from 50.5\% to 66.3\%), respectively. This indicates that the prior knowledge is necessary and the pretraining progress can bring significant improvement for AU detection.

Secondly, compared with the \textsl{w. GEE Emotion} and \textsl{w. GEE ImageNet}, \textsl{GTLE-Net} obtains the best performance(66.3\%), exceeding the other two pretraining methods about 6\% and 3.9\%. This proves the effectiveness and superiority of the prior knowledge captured by our facial expression similarity learning. 

Lastly, compared with \textsl{GTLE-Net}, \textsl{w. GEE fixed} obtains a lower average F1 of 63.5\%. It indicates that finetuning the global expression feature encoder is beneficial for extracting features which are more adaptive to the following \textsl{LAM} module. Thus finetuning operation is essential and efficient to improve the performance of our framework.

\textbf{Facial Components.}
To validate the effectiveness of the representations extracted by our local AU features module (\textsl{LAM}), we conduct three ablation experiments. Concretely, based on our full framework of \textsl{GTLE-Net}, we remove or replace some specific sub-modules and retrain the whole network, including:
\begin{itemize}
    \item [f) ]
    \textsl{w/o $E_m$\&$E_a$}: removing both the \textsl{$E_m$} and the \textsl{$E_a$};
    \item [g) ]
    \textsl{w/o $E_m$}: only removing \textsl{$E_m$};
    \item [h) ]
    \textsl{w/o $E_m$ \& w. CNN}: removing the \textsl{$E_m$} and replacing the \textsl{$E_a$} with a CNN block in the meanwhile. Particularly, the CNN block has similar number of parameters(around 16.6M) with our feature encoder($GEE$).
\end{itemize}

Note that the inclusion of \textsl{$E_a$} alone(\textsl{w/o $E_m$}) brings an improvement of 0.5\%(from 63.3\% to 63.8\%) in comparision to \textsl{w/o $E_m$\&$E_a$}. When the \textsl{$E_a$} is replaced by the CNN module(w/o $E_m$ \& w. CNN ), the performance declines by 0.1\%(from 63.8\% to 63.7\%).
Moreover, the F1-Score improves by 2.5\% when we further add the \textsl{$E_m$} module(\textsl{GTLE-Net}). These study results validate the effectiveness of our AU mask and local feature learning.



\begin{figure}[!t]
    \centering
    \includegraphics[ width=9cm]{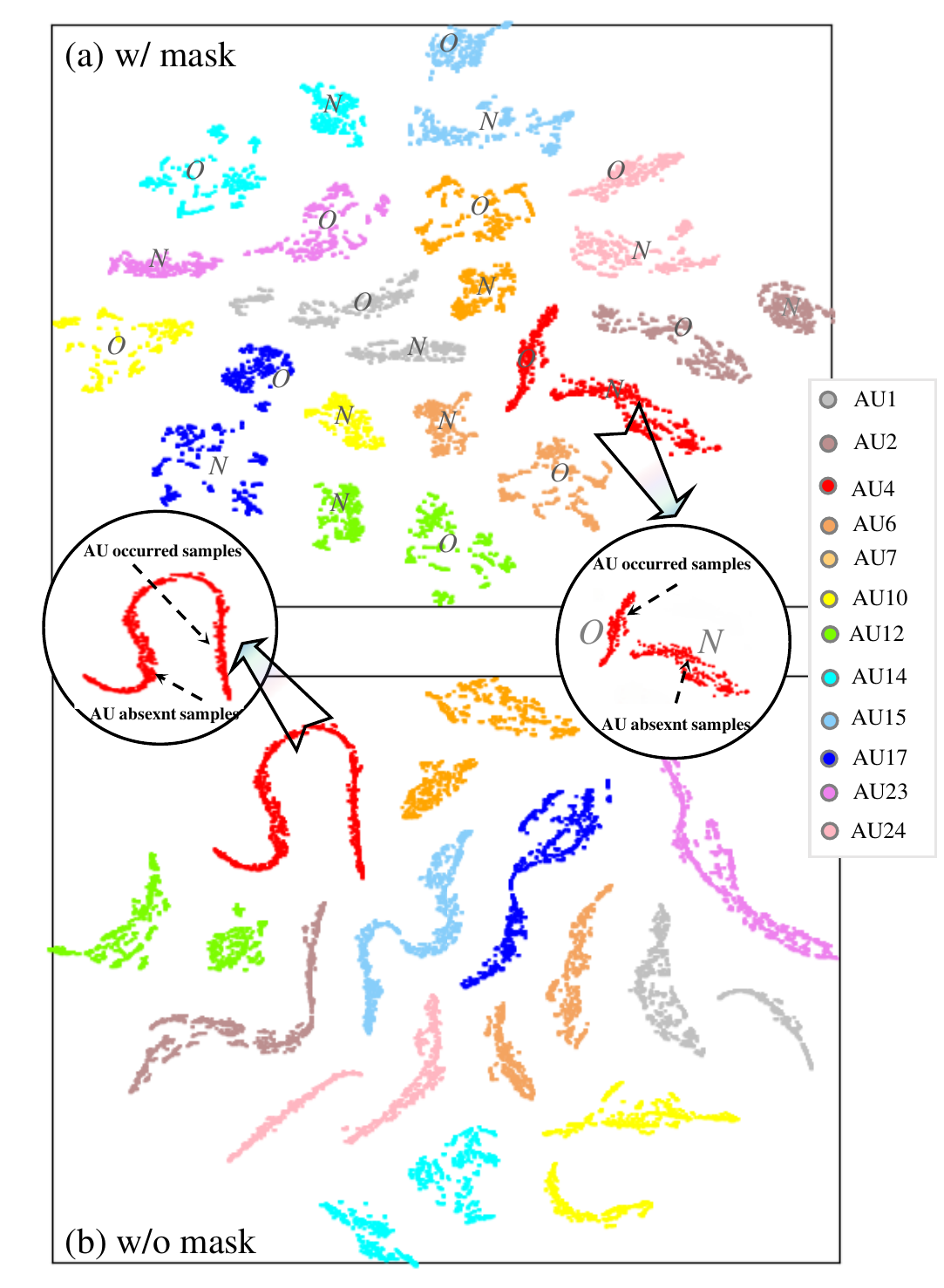}
    \caption{t-SNE visualization of the AU embeddings generated by our framework w/ or w/o the mask feature extractor($E_m$) module on the BP4D dataset. The AU embeddings w/ mask gather into two clusters of AU occurred($O$) and absent($N$) samples.}
    \label{fig:tsne}
\end{figure}

\subsection{Visual Analysis}
We also visualize the feature and mask results of our framework to further validate its effectiveness.

\textbf{Mask Attention Learning.} Due to its anatomical definition, each AU involves a group of muscle motions and is related to a specific region on the face. For instance, AU1 is inner brow raising and AU9 is nose wrinkle. Theoretically our framework should learn a regional mask activation map for each AU, corresponding to its muscle actions. 

Figure~\ref{fig:figure1} visualizes the attention mask maps of several 
examples from DISFA, BP4D and BP4D+, learned by our network. As shown in this figure, we can see that our method successfully captures a discriminative attention mask for these AUs. Taking AU4 (brow low) and AU17 (chin raise) as examples, the learned spatial masks are around the brows and 
chin, respectively, which demonstrates that our proposed AU mask extractor is effective for AU-related local regional attention learning.

From Figure~\ref{fig:mask visualization}, we can see that each AU's mask is closely related to its muscle region, even for different expressions such as happy, sad and fear.
We also notice that several different AUs may cover the same or similar area. This is attributed to that different AUs may share the same or similar muscle distributions. For instance, all of AU1, AU2 and AU4 are involved with the forehead, while all of AU14, AU15, AU17, AU23 and AU24 are closely related to the mouth area. Besides, considering the existence of potential mutual relations among different AUs, the masks learned merely supervised by AU labels may include other facial regions of correlated AUs. For instance, the masks of AU1, AU2, and AU4 not only cover the regions of the brows but also contain the mouth area.

\textbf{Mask Features Analysis.} To validate the importance of mask learning, we visualize the AU embeddings of the AU classifiers in the \textsl{GTLE-Net} and \textsl{w/o $E_m$}, respectively. We sample one thousand images and obtain the t-SNE visualization of their AU embeddings. 

As illustrated in Figure~\ref{fig:tsne}, each color refers to a class of AU presentations. Obviously, the presentations learned with mask(\textsl{GTLE-Net}) are more compact and have better distribution than that generated by model without mask(\textsl{w/o $E_m$}). Moreover, all the AU embeddings learned with mask(\textsl{GTLE-Net}) tend to gather into two clusters(occurred and absent samples). By contrast, the presentations generated without the mask extractor(\textsl{w/o $E_m$}) display like a long strip and are not divided into two clusters for the cases of AU2, AU4, AU15, AU17 and AU23. As a result, these AUs cannot be well distinguished. This validates that the proposed \textsl{LAM} of our framework is effective to capture discriminative local features that are beneficial to AU recognition.



\section{Discussion and Conclusion}
Inspired by the correlation between expression representation learning and AU detection, in consideration of the local properties of AUs, in this paper we propose a novel framework for AU detection by learning both expression representation similarity and facial attention. The expression representation learning aims at obtaining powerful and robust latent features that are sensitive to subtle expression transformations. On the other hand, to capture advanced local features concentrated on AU-related areas, we design a local AU feature module(\textsl{LAM}) to learn local attention maps indicating spatial significance and extract local AU representations.  

Extensive experimental comparisons with state-of-the-art methods validate that our proposed framework can measurably outperform all of them on the widely-used benchmark datasets (BP4D, DISFA, and BP4D+) for the AU detection task. Our ablation studies not only demonstrate the effectiveness of the prior knowledge from expression similarity learning, but also verifies the validity of our proposed local AU features module(\textsl{LAM}). Through visual analysis, we show that our method can produce highly AU-related attention maps that are sensitive to local facial muscle motions, and that our attention mechanism module can learn compact and discriminative AU embeddings. 
In addition, we demonstrate the generalization capability of our method through Cross-datasets validation experiments. 

While our method achieves superior results on three widely-used datasets, it does not work well for data in the wild. Besides, due to the limitations of AU annotation, it can just detect a part of AUs in FACS. Finally, our method does not explicitly consider the correlation between AUs, which may influence the eventual performance.

Besides AU detection, our method can be potentially extended for other facial analysis applications such as expression recognition and micro-expression detection. In the future, we plan to explore these research directions by extending the current framework.







\ifCLASSOPTIONcaptionsoff
  \newpage
\fi


\bibliographystyle{IEEEtran}
\bibliography{egbib}





\begin{IEEEbiography}[{\includegraphics[width=1in,height=1.25in,clip,keepaspectratio]{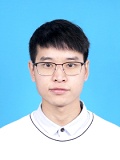}}]{Rudong An} received the B.S. degree and M.S. degree in college of Biomedical Engineering and Instrument Science from Zhejiang University, Hangzhou, China, in 2018 and 2021. He is currently an artificial intelligence researcher at Netease Fuxi AI Lab. His interests include deep learning, action unit detection, facial expression analysis and facial animation generation. 
\end{IEEEbiography}

\begin{IEEEbiography}[{\includegraphics[width=1in,height=1.25in,clip,keepaspectratio]{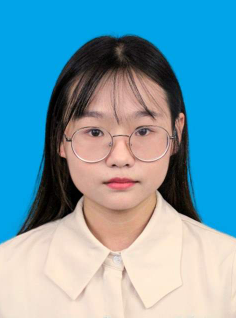}}]{Wei Zhang} received the B.E. degree in communication engineering from Nanjing University of Posts and Telecommunications, Jiangsu, China in 2017 and M.S. degree in electronic and information engineering from Zhejiang University, Zhejiang, China in 2020. She is currently a research scientist working with Netease Fuxi AI Lab, Hangzhou, China. Her current research interests include computer vision, expression embedding and facial affective analysis. 
\end{IEEEbiography}

\begin{IEEEbiography}[{\includegraphics[width=1in,height=1.25in,clip,keepaspectratio]{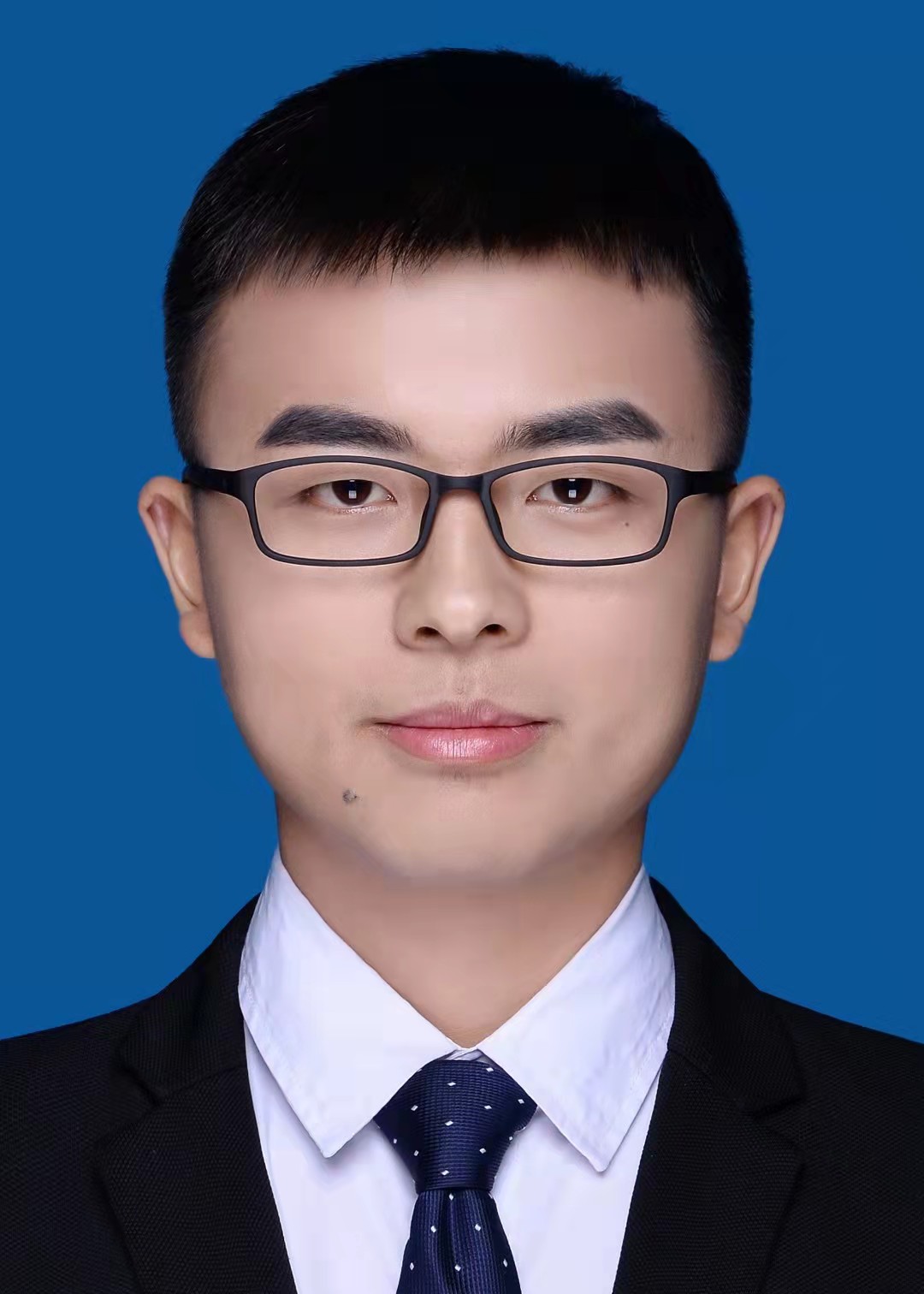}}]{Hao Zeng} received the B.E. degree in computer software engineering and the M.S. degree in computer science and technology from Huazhong University of Science and Technology, Hubei, China, in 2017 and 2020, respectively. He is currently a researcher in Netease Fuxi AI Lab, Hangzhou, China. His current research interests include deep learning, computer vision and face generation. 
\end{IEEEbiography}

\begin{IEEEbiography}[{\includegraphics[width=1in,height=1.25in,clip,keepaspectratio]{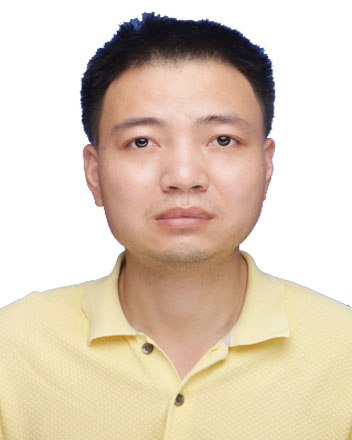}}]{Wei Chen}  received the B.S. degree in information engineering from the National University of Defense Technology, Changsha, China, and received M.S. degree in software engineering from Hebei University, Baoding, China. He is currently an engineer of Hebei Agricultural University, Baoding, China. His interests include software engineering, intelligent optimization algorithm, and deep learning. 
\end{IEEEbiography}

\begin{IEEEbiography}[{\includegraphics[width=1in]{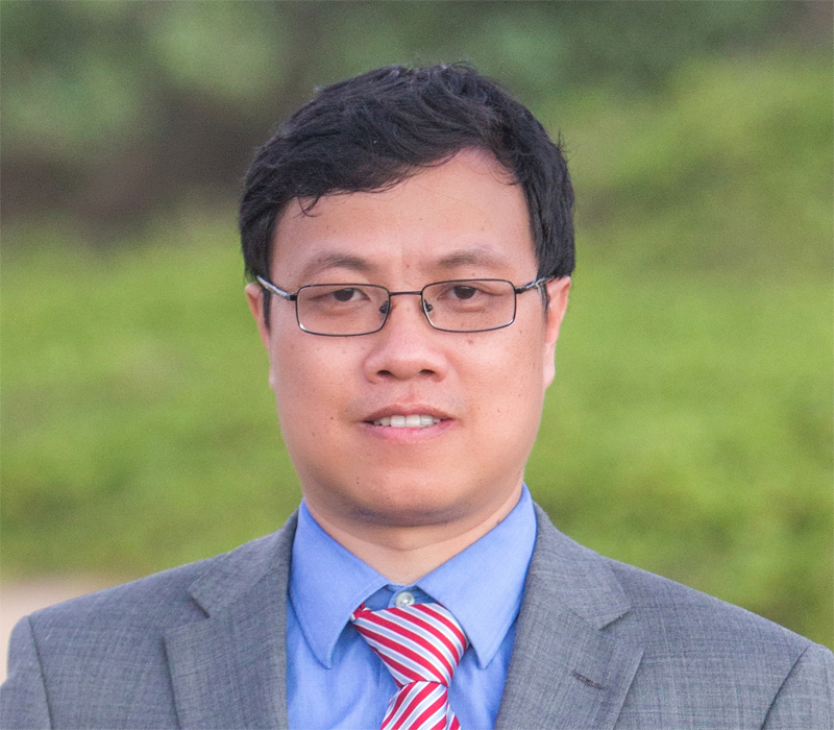}}]{Zhigang Deng}
is Moores Professor of Computer Science at University of Houston, Texas, USA. His research interests include computer graphics, computer animation, virtual humans, human computer conversation, and robotics. He earned his Ph.D. in Computer Science at the Department of Computer Science at the University of Southern California in 2006. Prior that, he also completed B.S. degree in Mathematics from Xiamen University (China), and M.S. in Computer Science from Peking University (China). Besides serving as the conference general/program co-chair for CASA 2014, SCA 2015, and MIG 2022, he has been an Associate Editor for IEEE Transactions on Visualization and Computer Graphics, Computer Graphics Forum, Computer Animation and Virtual Worlds Journal, etc. He is a distinguished member of ACM and a senior member of IEEE. 
\end{IEEEbiography}

\begin{IEEEbiography}[{\includegraphics[width=1in,height=1.25in,clip,keepaspectratio]{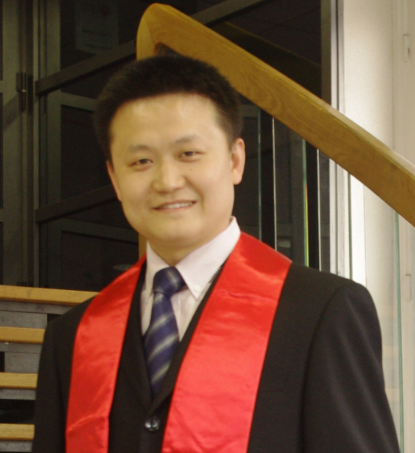}}]{Yu Ding} is currently an artificial intelligence expert, leading the virtual human group at Netease Fuxi AI Lab, Hangzhou, China. His research interests include deep learning, image and video processing, talking face generation, animation generation, facial expression recognition, multimodal computing, affective computing, nonverbal communication (face, gaze, and gesture), and embodied conversational agent. He received Ph.D. degree in Computer Science (2014) at Telecom Paristech in Paris (France), M.S. degree in Computer Science at Pierre and Marie Curie University (France), and B.S. degree in Automation at Xiamen University (China). 
\end{IEEEbiography}




\end{document}